\makeatletter
\def\input@path{{styles/}}
\makeatother
\documentclass[runningheads]{llncs}

\authorrunning{A. Attali et al.}
\institute{Department of Computer Science, University of Illinois\break 201 N. Goodwin Avenue, Urbana, IL 61801, USA}

\title{
A Framework for Guided Motion Planning
}

\usepackage{hyperref}
\usepackage[utf8]{inputenc}
\usepackage{authblk}
\usepackage[ruled, linesnumbered]{algorithm2e}
\usepackage{algorithmic}
\usepackage{marginnote}
\usepackage{amsfonts}
\usepackage{amsmath}
\usepackage{graphicx}
\usepackage{subcaption}
\usepackage[table,xcdraw,dvipsnames]{xcolor}

\usepackage{cite}
\usepackage[nameinlink]{cleveref}
\title{A Framework for Guided Motion Planning}

\author{Amnon Attali, Stav Ashur, Isaac Burton Love, Courtney McBeth, \break James Motes, Marco Morales, Nancy M. Amato}

\date{September 2022}

\newcommand{\R}{\mathbb{R}}
\newcommand{\C}{\mathcal{C}}
\newcommand{\X}{\mathcal{X}}
\newcommand{\GS}{\mathcal{S}}
\DeclareMathOperator*{\argmin}{argmin}

\begin{document}

\maketitle

\begin{abstract}
    Randomized sampling based algorithms are widely used in robot motion planning due to the problem's intractability, and are experimentally effective on a wide range of problem instances. Most variants bias their sampling using various heuristics related to the known underlying structure of the search space. In this work, we formalize the intuitive notion of \emph{guided search} by defining the concept of a \emph{guiding space}. This new language encapsulates many seemingly distinct prior methods under the same framework, and allows us to reason about guidance, a previously obscured core contribution of different algorithms. We suggest an information theoretic method to evaluate guidance, which experimentally matches intuition when tested on known algorithms in a variety of environments. The language and evaluation of guidance suggests improvements to existing methods, and allows for simple hybrid algorithms that combine guidance from multiple sources.
\end{abstract}

\section{Introduction}

A general paradigm in artificial intelligence is that one can solve provably worst case intractable search problems using heuristics that are effective in the average case, i.e., in most real world problems. This same principle holds true in robotics, where much of the difficulty stems from the geometry and the implicit nature of the search space, called the configuration space (C-space)~\cite{lw-apcfpapo-79}. The family of sampling based motion planning (SBMP) algorithms searches the continuous C-space using random sampling and local extensions in order to find valid paths.

Most such algorithms implicitly perform guided search. Early algorithms, such as Rapidly-exploring Random Trees (RRT) \cite{lavalle1998rapidly}, showed that exploration through Voronoi bias, i.e., choosing which node to expand with probability proportional to the volume of its Voronoi cell, is highly effective in the absence of arbitrarily narrow passages~\cite{klmr-rqprpp-95, hlm-ppecs-97}. Much work has centered around the narrow passage problem \cite{amato1998obprm, shi2014spark}. Due to the intractable nature of motion planning~\cite{r-cpmpg-79}, tangible improvements often came from making assumptions about the underlying space, such that the human engineer could encode case specific heuristics to bias the search. Some work, rather than improving runtime, focused attention on the types of solutions output by the algorithm, for example searching for paths with high clearance \cite{geraerts2007creating, was-maprm-99, ydlta-umaprm-14, ghk-prpfowmabsa-99, HollemanK00}.

In this work we make this previously implicit notion of guidance explicit by formally defining the concepts of \emph{guiding space} and \emph{guided search}. We define a guiding space as an auxiliary space that helps estimate the value of exploring each configuration. This value helps our proposed general guided search algorithm determine its next exploration step. Our definition of guidance is naturally hierarchical, as any search space (e.g., C-space) can itself be used as a guiding space. Intuitively, guidance is the bias introduced into the exploration process. Our algorithm is similar to A*, which performs heuristic guided search in \emph{discrete} spaces, but does so in C-space. 

In formally discussing guiding spaces we make the source of guidance explicit. Doing so allows us to identify guidance-generating components in many seemingly distinct prior methods, which fit under three main subcategories; robot modification, environment modification, and experience-based guidance. The perspective of guiding spaces helps isolate conceptual contributions of different algorithms and consequently yields simpler generalized implementations. These new implementations are more easily composed into algorithms that take advantage of the strengths of independent methods. Another benefit of framing SBMP algorithms in this common language is that it enables us to define a new method for evaluation that focuses on the quality of guidance.

We approach the task of evaluating guidance from an information theoretic perspective, whereby the procedure for computing guidance conveys information to the search process. We test our metric experimentally on a range of algorithms and environments. The results match intuition for the suitability of each method to features in certain C-spaces. Moreover, the metric captures properties otherwise obscured by the traditional metrics of runtime or number of samples, such as sensitivity to local environment features. Measuring quality of guidance locally gives insight into the value of combining guiding spaces, which we demonstrate with a hybrid guiding space that outperforms its individual parts.

\section{Preliminaries}

A motion planning problem consists of the tuple $M=(r,E,s,t)$, for robot $r$, environment $E$, and task $(s,t)$, where the objective is to find a (possibly shortest) valid path $P_{s,t}$ from starting configuration $s$ to goal configuration $t$. Note that $M$ is the single-query variant of the motion planning problem, and we will touch on the multi-query version later in this section. Together, $(r,E)$ describe the configuration space $\C$, a continuous space consisting of all possible configurations of $r$ in $E$. This space is implicitly partitioned into two parts, $C_{free}$ and $C_{obs}$. $C_{free}$ is the set of all valid configurations, and $C_{obs} = \C \setminus \C_{free}$. 

Incremental SBMP methods, such as RRT, explore $\C$ to solve $M$ by building a \emph{search tree} $T$. They do so by iteratively picking an existing node $v \in T$, and then expanding the tree by adding a new node as a child of $v$. We refer to these two operations as \emph{node selection}, which is performed by a sampler, and \emph{node expansion}, which is performed by a local planner. We refer to the set of points reachable by the local planner as $N_\C(v)$ for $v\in T$, the neighborhood of $v$. Usually the initial tree contains only the start node $s$ and the algorithm terminates when a configuration sufficiently close to $t$ is added to the tree.


Note that in this work we depart from the classical partition into tree and graph-based (roadmap)  algorithms (e.g., PRM \cite{KavrakiSLO96}). Most variants of graph-based algorithms attempt to solve the multi-query version of the problem. Our perspective is that such methods take advantage of a powerful prior; that future motion planning instances will be in the same C-space $\C$ but have different tasks. As such, preprocessing $\C$ to produce a data structure (the graph) which guides future motion planning queries by reducing them to path planning is highly effective. This perspective motivates our definition of guided motion planning: a tree based search algorithm that uses prior knowledge to generate a guiding space which informs exploration.

\section{Defining Guided Planning}
\label{sec:definition}

\begin{figure}[t]
    \centering
    \begin{subfigure}[t]{0.23\textwidth}
         \includegraphics[width=\columnwidth, page = 4]{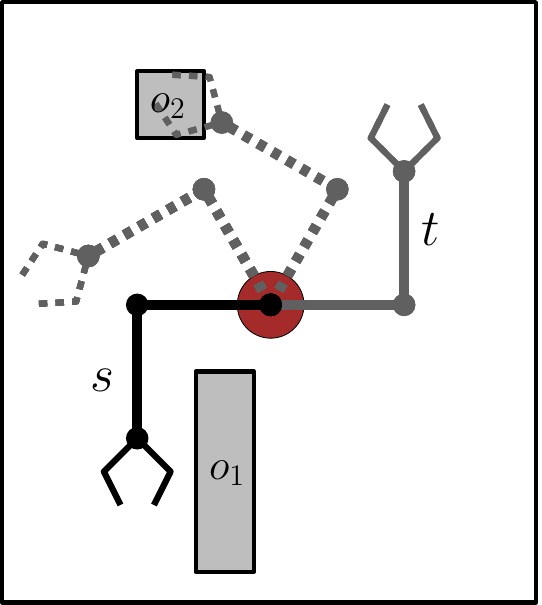}
         \subcaption{}
         \label{fig:guiding space usage: workspace}
    \end{subfigure}
        \hfill
    \begin{subfigure}[t]{0.23\textwidth}
         \includegraphics[width=\columnwidth, page = 3]{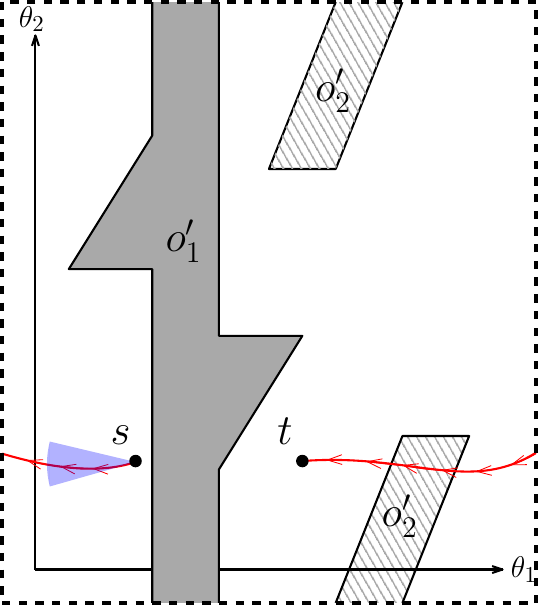}
         \subcaption{}
         \label{fig:guiding space usage: C-space}
    \end{subfigure}
        \hfill
    \begin{subfigure}[t]{0.23\textwidth}
         \includegraphics[width=\columnwidth, page = 5]{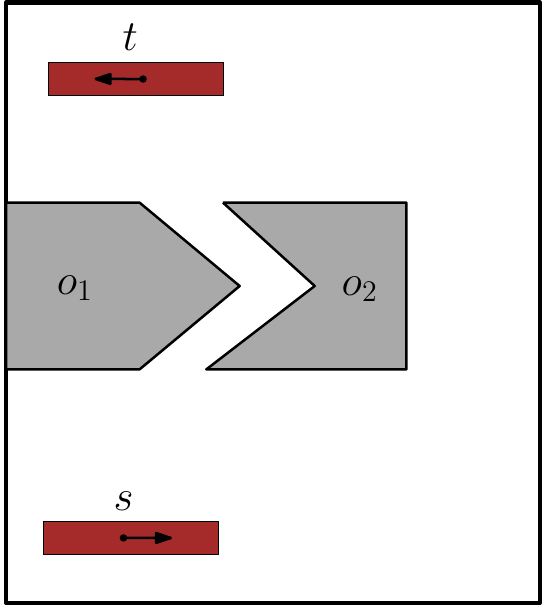}
         \subcaption{}
         \label{fig:guiding space usage 2: workspace}
    \end{subfigure}
        \hfill
    \begin{subfigure}[t]{0.23\textwidth}
         \includegraphics[width=\columnwidth, page = 6]{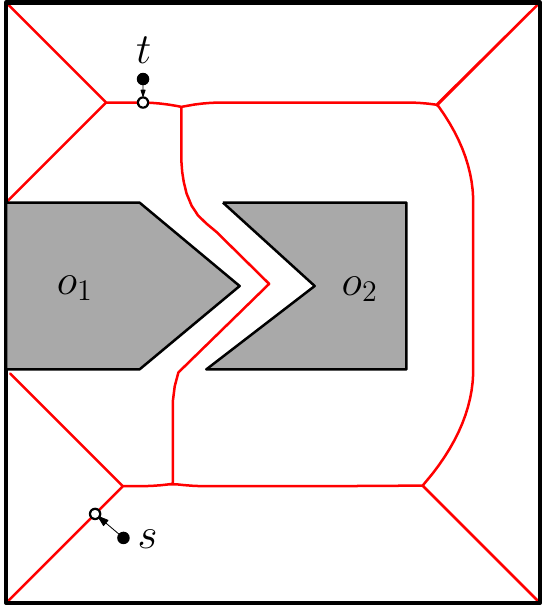}
         \subcaption{}
         \label{fig:guiding space usage 2: cspace}
    \end{subfigure}
    \caption{An illustration of guiding spaces and their usage. (a) depicts a 2D workspace with a fixed-base 2-joint manipulator and 2 obstacles, alongside two configurations $s$ and $t$ and  two intermediates (in dashed gray) on the $(s,t)$-path found in the guiding space (b). We observe that one intermediate is invalid. In (b) we see the lazy planning guiding space for the environment (a) without obstacle $o_2$. A red curve shows a path between the current and target configurations shown in (a), and a blue wedge shows the resulting sampling bias. In (c) we see another 2D environment with a 3 DoF $(x,y,\theta)$ rectangular robot, and in (d) we see a guiding space, the workspace medial-axis, and the projections of the configurations $s$ and $t$ shown in (c) onto the medial-axis. Again, note that the shortest $(s,t)$ path in (d) is not valid in (c).}
    \label{fig:guiding space illustration}
    
\end{figure}

In this section, we define the main components of guided motion planning: namely the guiding space, a mapping which relates C-space to the guiding space, and an algorithm that uses both to plan. We provide some simple intuitive examples followed by a categorization of guiding spaces in the next section.

At a high level we think of guidance as the knowledge distilled from past experience to be used in future motion planning. Thus, in the setting of guided motion planning, we assume there exists an unknown but fixed distribution over robots, environments, and tasks. Guided search consists of an optional \emph{offline} training phase, during which we learn from this distribution, followed by an \emph{online} phase, in which a specific motion planning problem is sampled from the distribution and solved. We will also show that a \textit{search} space can be used as a guiding space, making the definition naturally hierarchical since that search space can, in turn, have guidance.

Note, we are generally not concerned with time complexity during the offline phase, and that many algorithms skip it entirely. For example, RRT does not ``learn'', and performs the same procedure regardless of C-space. More commonly, after making assumptions about the C-space distribution, the human engineer encodes their prior knowledge into a suitable procedure for computing guidance, thus replacing the training phase. For example, knowing that some obstacles in the environment are fixed might motivate precomputing collision with those obstacles \cite{jaillet2004prm} (where PRM is an extreme version of this). Alternatively, knowing that some degree of freedom has negligible impact on $\C$ might motivate planning with only the other dimensions \cite{lozano1987simple, xa-kbprm-04}.

We now define the individual components of guided motion planning to make explicit where and how guidance is used.

\subsection{Guided Planning Definition}
We define a \emph{search space} $\X$ as a state space (set) with a distance function $d_\X: \X\times \X \rightarrow \R_+$. In contrast, we define a \emph{guiding space} $\GS$ as a state space which assigns a value to each state, namely there exists some function $h_\GS : \GS \rightarrow \R_+$, which we call the \emph{heuristic}. Note that a C-space $\C$ is a search space and that every search space can be viewed as a guiding space, i.e., $h_t(x) = d(x, t)$ for $x\in \C$. In other words, distance to some goal state $t$ can be used as a heuristic.

A \emph{guiding space projection} $P_T : \X \times \X \rightarrow \GS$ is a mapping from pairs of states (tasks) in some search space $\X$ to states of a guiding space $\GS$. This mapping is conditioned on a search tree $T$, which intuitively represents the known information regarding $\C_{free}$ and $\C_{obs}$. By relating tasks in $\X$ to states in $\GS$ the projection allows a guided search algorithm to use the heuristic during search. Often for convenience we will drop the second term to the projection, namely $P_T(x) = P_T(x, t)$. The offline training phase of guided motion planning produces a procedure $P^M_T$ for computing a projection for a new motion planning problem by training on samples $M \sim \mathcal{D}$ for some distribution $\mathcal{D}$ over robots, environments, and tasks.

A \emph{guided motion planning problem} $M_P=(r,E,s,t,P)$ is a motion planning problem augmented with a projection $P$. We note that the image of $P$ describes a (potentially implicit) guiding space $\GS$ and corresponding $h$.

Consider the motion planning problem $M=(r,E,s,t)$ depicted in Figure~\ref{fig:guiding space usage: workspace}, where a two joint planar manipulator $r$ must navigate from state $s$ to $t$ while avoiding obstacles $E = \{o_1, o_2\}$. Now consider the related C-space in Figure~\ref{fig:guiding space usage: C-space} which omits obstacle $o_2$, $\C'=(r,E'), E'=\{o_1\}$. Let $\GS$ be the guiding space whose heuristic is the distance to $t$ in $\C'$, i.e., $h_\GS (x) = d_{\C'}(x,t)$. In this example the guiding space projection is $P_T((x, t)\in \C \times \C) = x \in \GS_t$, where $h_{\GS_t}(x) = d_{\C'}(x, t)$. Notice that it does not depend on the search tree, and uses distances in $\C'$ as heuristics (to guide search in $\C$). In addition, while the projection may seem like the identity mapping $x\mapsto x$, it is implicitly modifying the validity of $x$ since configurations which collide only with obstacle $o_2$ in $\C$ are valid in $\C'$. Despite colliding with obstacle $o_2$, the path highlighted in red still contains useful information for computing paths in $\C$. Not only is the length of this path a good estimate for the true shortest path length, but also at certain configurations it contains locally near-optimal controls. The key insight for guidance is that not only does this path contain useful information, but also it is easier to compute than the true optimal solution, since it corresponds to a simpler search problem.

\begin{algorithm}[t]
    \SetAlgoLined
    \SetKwInOut{Input}{input}
    \Input{$M_P = (r, E, s, t, P)$}
    Initialize tree $T=\{s\}$\\
    \While{$\neg$ found goal}{
        select node: $x_{min} = \argmin_{v\in T} h(P_T(v))$\\
        expand node: $x_{new} = \argmin_{v\in N_\C(v)} h(P_T(v))$\\
        $is\_valid = (x_{min}, x_{new}) \subset \C_{free}$ \\
        if $is\_valid$: add $x_{new}$ to $T$ as child of $x_{min}$ \\
        else: add $(x_{min}, x_{new})$ to $T.failed\_expansions$
    }
    \caption{Guided Search}
    \label{gs_alg}
\end{algorithm}

\subsection{Guided Search}

Given a search tree $T$, a baseline exploration algorithm might explore the space by selecting an existing state on the tree uniformly at random and then expanding from it according to some local planner. Unfortunately this approach has poor exploration properties\cite{lavalle1998rapidly}, as it is heavily biased towards regions that have already been explored. Discrete search avoids this problem as newly explored states are checked against a cached set of visited states.

Algorithm~\ref{gs_alg} shows an example procedure for the online phase of guided search for solving a problem $M_P$. In each iteration the algorithm maps nodes on the tree through the projection $P_T$ then uses $h$ to determine the next node to explore, i.e., by selecting the node with minimum $h$. In general, we can think of the projection as providing guidance by computing a distribution over the nodes of the tree from which we can sample to select how to expand. The tree then grows according to the neighborhood defined by the local planner. 

This parallels the well known heuristic guided discrete search algorithm A*, where the main difference is that instead of maintaining an explicit ordered frontier for search, we allow the guiding space to update its distribution over $T$ at each iteration. This distinction is crucial - each new sample provides information regarding which regions of space to explore such that no fixed ordering of states (e.g., A*) is an effective exploration strategy.

\subsection{Simple Examples of Guiding Spaces} 
\label{sec:simple_examples}

One guiding space for $\C$ is itself, $\GS = \C$, where $P_T(x) = x$ is the identity and $h = d_{\C}(x,t)$ is the true distance to the goal configuration.
This highlights the general setting in which a search space is used as a guiding space, in which, for guidance to be useful, we expect that solving tasks in $\GS$ should be easier than in $\C$.
Similarly, $\GS = \{0\}$ is a guiding space with a trivial projection $P$ and uninformed $h$.
Such trivial projections correspond to fixed sampling distributions over $T$. 

We observe a natural trade-off between guidance quality and computation difficulty - computing solutions when $\GS = \C$ is difficult (the original problem) but they provide perfect guidance, whereas when $\GS = \{0\}$ we can compute $h$ in constant time but get uninformed guidance.

The classic implementation of RRT samples uniformly in $\C$ to select which node on the tree to expand. This is not equivalent to sampling a node uniformly at random from the tree, and is known to  guide exploration towards unexplored regions via Voronoi bias. While in practice the guiding space is implicit in the sampling and nearest neighbor procedure, we can compute it explicitly through the Voronoi decomposition, where the value $h$ of each node on the tree is (inversely) proportional to the volume of its corresponding cell. Notice how the guiding space depends on the search tree, as the nodes dictate the Voronoi cells. 

As mentioned previously, a PRM in configuration space is a guiding space. If the distribution $\mathcal{D}$ contains a single robot and environment, but is stochastic over tasks, then precomputing a discrete roadmap in $\C$ is a common approach for answering multiple queries. The projection $P$ is often taken to be the function which returns the nearest neighbor on the graph. For any task $(s,t)$ and vertex $v = P(s)$, the heuristic $h_\GS (v)$ is the length of the shortest path on the graph between $v$ and $P(t)$. This length is an approximation to the optimal solution for that task and can sometimes provide imperfect guidance; for example not every point in space might be locally navigable to a node on the graph. 

One of the most commonly used forms of guidance comes from the workspace (see Figure \ref{fig:guiding space usage 2: cspace}). Planning for a point robot in the low-dimensional workspace is relatively easy, and while not every workspace path is feasible in C-space, every C-space path corresponds to some feasible path in workspace. In other words, in the context of A*, we can think of distance in workspace as an admissible heuristic for distance in C-space (with some possible scaling). In the next section we describe how workspace guidance methods can be thought of as using plans for a modified version of the robot to guide planning in the original C-space. 


\section{Categories of Guidance}
\label{sec:prior_work}

We separate guiding space methods into three main categories: robot modification, environment modification and experience-based guidance. We find these subcategories natural and comprehensive, as the entire motion planning problem is encapsulated in the structure of the configuration space, which can be learned from experience, or is induced by its robot and environment constraints.

In this section, as we define the categories of guidance, we discuss how existing work fits into each category. Note that much of the work cited below does not explicitly describe guidance as a contribution, yet here we identify that they do implicitly perform guided search. 



\subsection{Robot Modification}

Recalling that a configuration space is the product of a robot $r$ and environment $E$, $\C = (r, E)$, we define robot modification as producing a guiding space $\GS = (r', E)$ for some new robot $r'$. In this paradigm, the projection function $P$ maps configurations in the first robot's C-space to configurations in the second. 

Given such a mapping $P$, $h$ generally computes paths between configurations in $\GS$. Notice that the translation of a path back into $\C$ is often costly and may be one-to-many (e.g., inverse-kinematics), but isn't necessary because we are only interested in the \emph{value} of each state. It is then the job of the guided search algorithm to translate this value into feasible paths in $\C$. 

There are many natural ways one might modify a robot description to obtain a closely related configuration space. A useful form of robot modification approximates the robot with simple geometries, such as bounding volumes, to speed up collision detection \cite{khmsz-ecdubvhkdop-98}. One of the most natural projections is to define $\GS$ as workspace itself, ignoring the robot entirely and treating it as a fully controllable point robot in the environment \cite{holleman2000framework, DennySBA16}. Some prior work ignores kinodynamic constraints on the robot \cite{PlakuKV07}, thereby yielding a guiding space where the robot is locally controllable. For chain-link robots, such as manipulators, a common guiding space involves focusing on a subset of the degrees of freedom \cite{lozano1987simple}. Such methods often use heuristics in deciding which subsets to consider, e.g. focus on the first joints of a manipulator. Finally, some work attempts to learn a subspace of the robot's degrees of freedom, e.g., using statistical projections such as PCA \cite{MahoneyBJ10}.

In general, we may think of robot modification methods as performing transfer learning in which examples of one robot performing tasks are used to inform plans for a second robot \cite{devin2017learning}. In the absence of a ground truth robot model, one can be learned from data \cite{bocsi2013alignment}. 

All the above robot modification methods produce a search space as opposed to a guiding space. For example, workspace guidance methods may build a sparse graph in workspace, such as the medial axis \cite{DennySBA16}. Therefore there is often a secondary guided search that occurs in the modified robot space, e.g., the medial axis graph. This both highlights the hierarchical nature of guidance and helps isolate contributions - the medial axis is a useful structure for guiding search in workspace, and workspace is useful for guiding search in C-space.

\subsection{Environment Modification}
Environment modification refers to guiding spaces which map $\C=(r, E)$ to $\GS=(r, E')$. Unlike in robot modification, the associated projection $P$ is often simpler to compute, and translation of paths in $\GS$ back to $\C$ requires relatively little work. In fact, in all work mentioned below, $P$ is nothing but the identity function, which only changes the validity of some points in $\C$ to produce $\GS$.

Narrow passages are a known bottleneck for randomized sampling algorithms. As such, some work modifies obstacles in an attempt to widen passages \cite{BayazitXA05, VonasekPK19}. Lazy planning methods \cite{BohlinK00, Hauser15} also perform a type of environment modification by ignoring constraints during initial planning, and incorporating them later as needed. When removing constraints, paths in $\GS$ are not necessarily valid in $\C$, and thus often such methods iteratively or hierarchically compute a guiding space (e.g., the initial lazy graph), use $h$ to produce samples near the unvalidated path, and then repeat with a new guiding space informed by the recently acquired samples (e.g., remove edges found to be invalid). 

In contrast to the removal of environment constraints, adding constraints is less common, but can be done to simplify geometries and therefore collision detection \cite{GhoshTMRA16}. Moreover, paths found when adding constraints are always valid in the original search space. In other words, if we take $h$ to be the cost of paths in the modified space, constraint removal provides optimistic lower bounds whereas constraint addition produces conservative upper bounds.

\subsection{Experience-Based Guidance}

The most common version of experience-based guidance assumes the dataset contains a variety of environments all for the same robot. Consequently, database queries often consist of evaluating the degree to which paths in the dataset violate discovered constraints in the current motion planning instance \cite{berenson2012robot}. While some methods only query using the task endpoints \cite{coleman2015experience}, more recent approaches propose to query (or filter) the database using local task features derived from workspace occupancy \cite{chamzas2021learning}. Another approach projects trajectories onto a lower dimensional abstract space via auto-encoding \cite{ichter2019robot} or metrics like dynamic time warping \cite{aine2015learning}, which improves database query. Some methods perform multiple queries, where the guiding space effectively becomes the composite space of multiple independent motion planning problems from the dataset, e.g., through gaussian mixture models \cite{chamzas2021learning}. 

At the extreme end we have learning based methods which train a neural network from a dataset of experience to provide guidance, effectively combining all past solutions through the weights of the neurons of a neural network. In encoder-decoder models \cite{qureshi2019motion} we may think of the encoder as $P$, which maps the input configuration space onto an abstract guiding space, and the decoder as $h$, which solves the task in the abstract space to provide guidance, say through predicting cost-to-go \cite{bhardwaj2017learning}. Many learning based methods have poor generalization due to dependence on problem specific features, e.g., top down 2D occupancy maps \cite{wang2020neural}. Those with good generalization often scale poorly by learning an equivalent to cost-to-go \cite{bhardwaj2017learning}, which even in a discrete space corresponds to computing or storing a prohibitively expensive all pairs shortest path.

In all learning based approaches a major challenge is gathering appropriate data, leading to methods which simplify the problem by learning local guidance through learned local planners \cite{faust2018prm}. Some work \cite{attali2022discrete} proposes to gather appropriate data for training by using SBMP.

\section{Evaluation metric}
\label{sec:metric}

Often we evaluate sampling based motion planning methods using algorithm level holistic metrics such as running time or number of samples. While minimizing such metrics is a reasonable end goal, they can be sensitive to implementation details, and focusing on them can obscure which aspect of a proposed contribution is responsible for the results. In the context of guided motion planning, we now present a new method to evaluate quality of guidance. We also note that prior work has considered alternative metrics for evaluating samples in motion planning, but in the context of giving a roadmap certain properties rather than guidance for a tree-based method~\cite{m-mfsbmp-07, burns2003information}.

Tree-based search algorithms first select a node on the tree then expand the selected node to a new child. We focus on evaluating \emph{node selection} as opposed to node expansion, and note that evaluating node expansion can be done by adding all possible (or a random sampling of) children to the tree, and using the evaluation of the best child according to our node selection evaluation.

\subsection{Sampling Efficiency}

\begin{figure}[t]
    \centering
    \begin{subfigure}[t]{0.23\textwidth}
         \includegraphics[width=\columnwidth, page = 5]{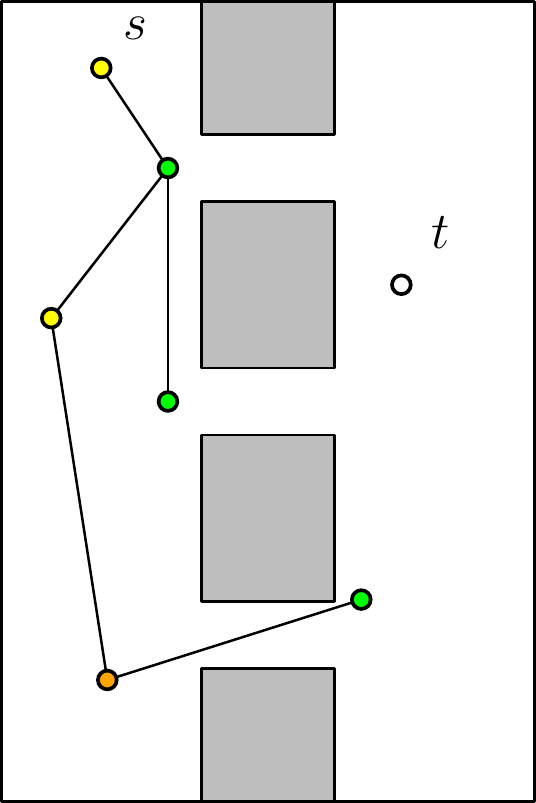}
         \subcaption{}
         \label{fig:se_efficiency2}
    \end{subfigure}
        \hfill
    \begin{subfigure}[t]{0.23\textwidth}
         \includegraphics[width=\columnwidth, page = 2]{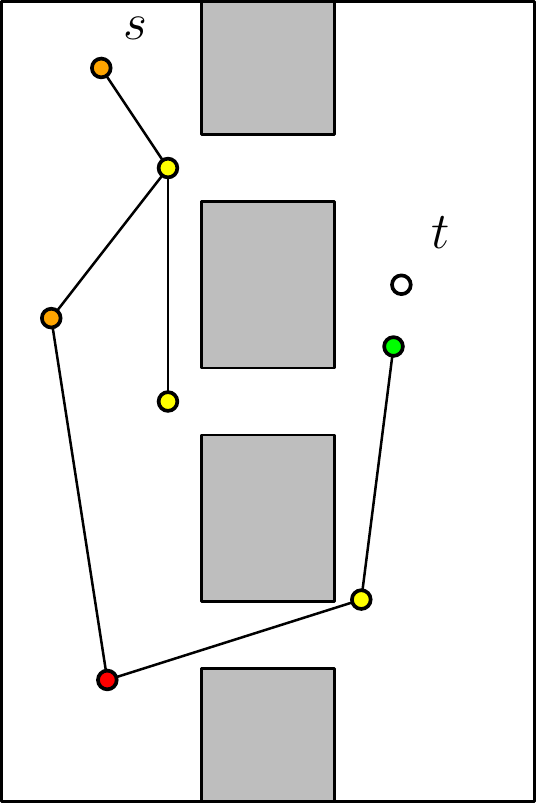}
         \subcaption{}
         \label{fig:se_efficiency1}
    \end{subfigure}
        \hfill
    \begin{subfigure}[t]{0.23\textwidth}
         \includegraphics[width=\columnwidth, page = 3]{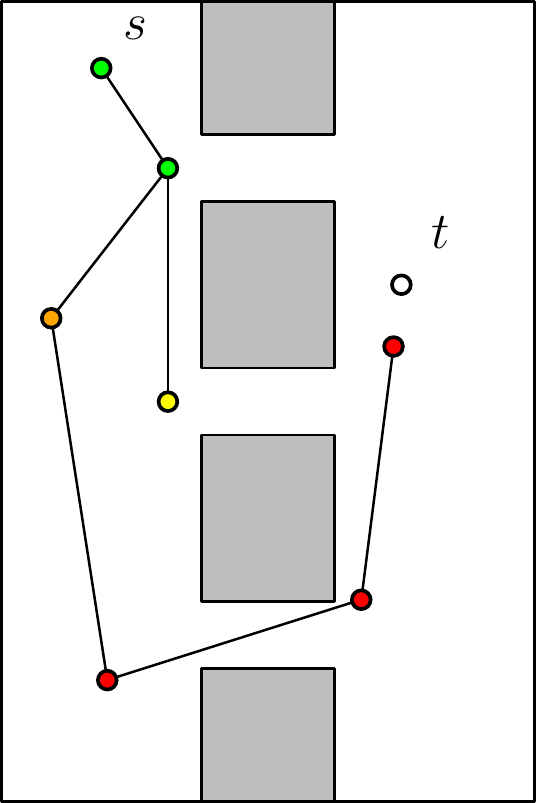}
         \subcaption{}
         \label{fig:se_optimality1}
    \end{subfigure}
        \hfill
    \begin{subfigure}[t]{0.23\textwidth}
         \includegraphics[width=\columnwidth, page = 4]{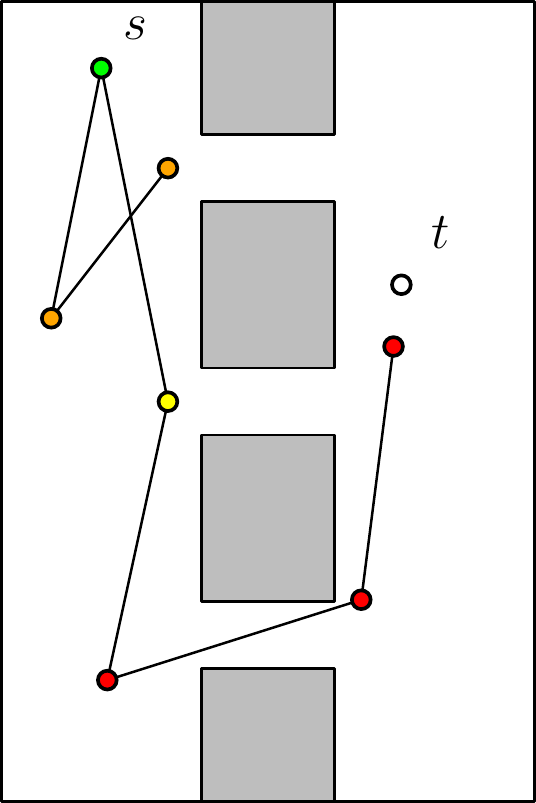}
         \subcaption{}
         \label{fig:se_optimality2}
    \end{subfigure}

    \caption{A simple 2D example of the components of sampling efficiency: \textit{efficiency} and \textit{(sub)-optimality}. The nodes are colored on a green (best) to red (worst) scale. (a) and (b) show the efficiency score of each node, which corresponds to the work remaining. (c) and (d) show the suboptimality scores of the nodes, which captures the deviation of each node from the optimal path. Notice that the value of a node is \textit{relative} to the other nodes, e.g., adding a new sample from (a) to (b) changes the value of all others. Notice that suboptimality depends on the tree structure, e.g., the rewiring between (c) and (d) changes the values of the nodes. Finally, despite depicting identical trees, (b) and (c) show how efficiency and suboptimality produce very different distributions over the nodes.}
    \label{fig:se_gradient}
    
\end{figure}

We propose to evaluate guidance by comparing the distribution of samples produced by the guiding space to some target distribution, namely one derived from an optimal sampling distribution with oracle access. 


\begin{definition}[Sampling efficiency]
Given a tree $T$, a target sampling distribution over the tree nodes $Q$, and an empirical sampling distribution $P$, the sampling efficiency of $P$ is defined as the information gain of using the target $Q$ instead of $P$, equivalently described as the Kullback-Liebler divergence between the two distributions, 
$$SE_Q(P) = D_{KL}(P || Q) = \sum_{v\in T} P(v) \log \frac{P(v)}{Q(v)}$$
\end{definition}

When $P$ and $Q$ are very similar $SE_Q(P)$ is close to zero, and grows as they grow far apart. In the case where our empirical distribution is deterministic, $P(u) = 1, P(v) = 0, v\neq u$, we have, $$SE_Q(P) = \sum_v P(v)\log \frac{P(v)}{Q(v)} = \log \frac{1}{Q(u)} = - \log Q(u),$$ reducing to the negative log-likelihood of the empirical sample under the target distribution.

Intuitively, we aim to measure the difference between these two distributions. While the KL-Divergence provides such a measure, it can be dominated by small regions in space where the two distributions have the largest difference (e.g., when one distribution assigns zero probability to a region in which the other is non-zero). In sampling based motion planning we often aren't worried about a few bad samples, so long as good ones are made with high enough probability so the bounded and symmetric Jensen-Shannon Divergence metric provides a smoother comparison, i.e., $$D_{JS}(P||Q) = \frac{1}{2} \left [D_{KL}(P || M) + D_{KL}(Q || M)\right ],$$ 
for mixture distribution $M = \frac{P+Q}{2}$. $D_{JS}$ is therefore useful as it is less influenced by outliers in the distributions. Alternatively we can assume the target $Q$ assigns a minimum nonzero probability to every node, $Q(v) \geq \epsilon$ for all $v \in T$, and therefore $SE_Q(P) \leq \log \frac{1}{\epsilon}$. In this way $\log \frac{1}{\epsilon}$ captures the maximum penalty we associate with poor sampling decisions made by the guiding space.

In shortest path problems, we are interested in (efficiently) computing a boundedly optimal $(s,t)$ path in $\C$. Let $\delta_v$ be the suboptimality of $v \in T$, $\delta_v = d_T(s,v) + d_\C(v,t) - d_\C(s,t) > 0$, where $d_T$ represents the distance between nodes on the tree and $d_\C$ is the true shortest path between configurations in $\C$. Now let $\tau_v$ be the work remaining for a search starting at $v$, $\tau_v = d_\C(v,t)$. Figure~\ref{fig:se_gradient} shows an intuitive example of the relationship between these quantities. We propose to define $Q$ with respect to $\tau_v$ and $\delta_v$:
$$Q_T(v) \propto \exp(-\frac{\delta_v}{\delta})\exp(-\frac{\tau_v}{\tau}) = \frac{e^{-(\delta_v/\delta + \tau_v/\tau)}}{Z},$$ $$Z = \sum_x e^{-(\delta_x/\delta + \tau_x/\tau)}$$
i.e., $Q_T(v) = softmin(\delta_v/\delta + \tau_v/\tau)$ for temperature parameters $\delta, \tau$ which naturally balance the relative importance of optimality and efficiency. The smaller these parameters the steeper the distribution, assigning higher probability to samples with low $\delta_v$ and $\tau_v$, and more heavily penalizing poor samples. 

We can now define the smoothed $\bar Q_T \geq \epsilon$ as $$\bar Q_T (v) \propto \exp(-(\delta_v/\delta+\tau_v/\tau)) + \gamma,$$ where 

$$\min_u \frac{e^{-(\delta_u/\delta + \tau_u/\tau)} + \gamma}{Z + \gamma|T|} \geq \epsilon \implies \gamma \geq \max_u \frac{Z (\epsilon - Q_T(u))}{1-\epsilon|T|}$$

We refer the reader to \cite{aaa-mgmpdp-24} for an in depth discussion of a metric to evaluate the quality of a sample for ST-connectivity problems in $C$-space; where measuring progress for the decision problem is not as straightforward as computing the cost-to-go.

\begin{figure*}[t]
\centering
\subfloat[SimplePassage]{\label{fig:task1}\includegraphics[width=0.23\linewidth]{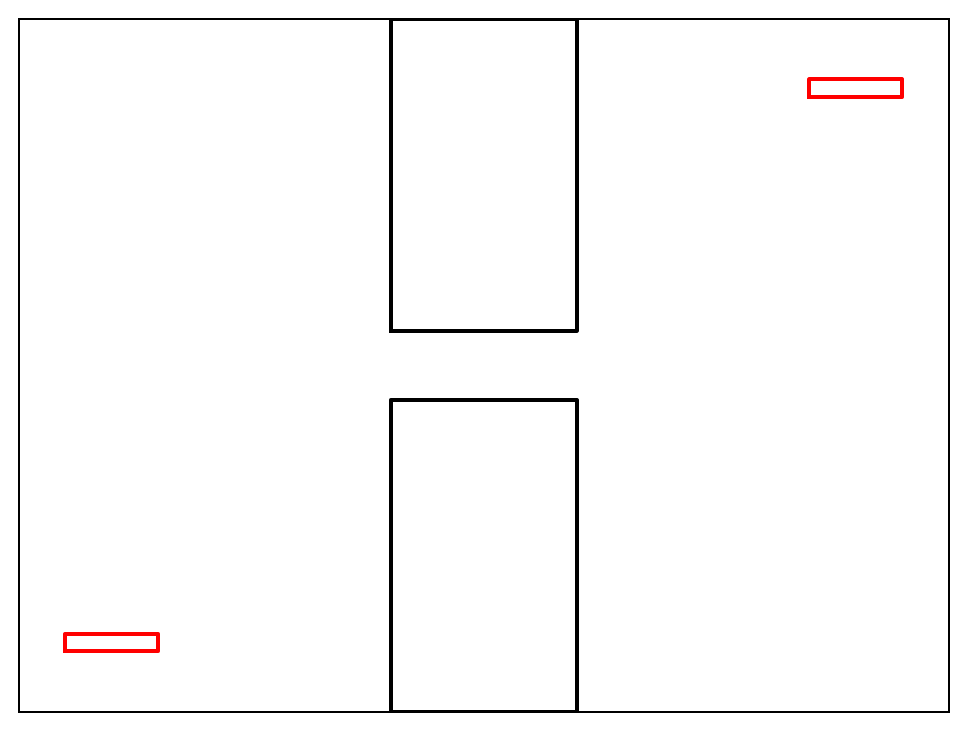}}
\hfill
\subfloat[Cup]{
\label{fig:task2}\includegraphics[width=0.23\linewidth]{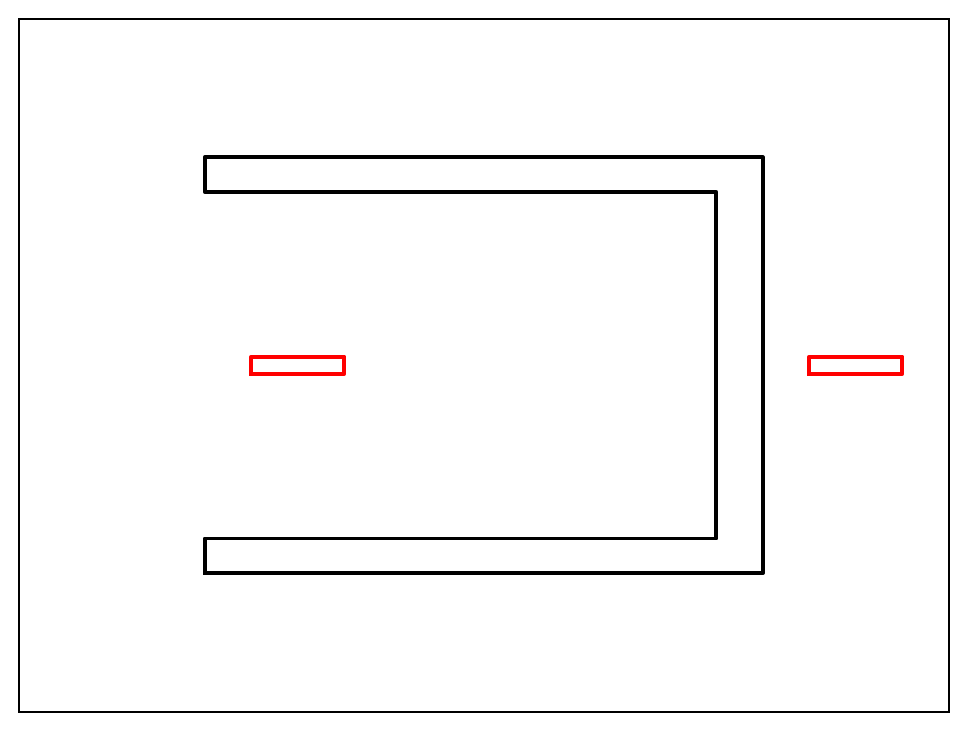}}
\hfill
\subfloat[Trap]{\label{fig:task3}\includegraphics[width=0.23\linewidth]{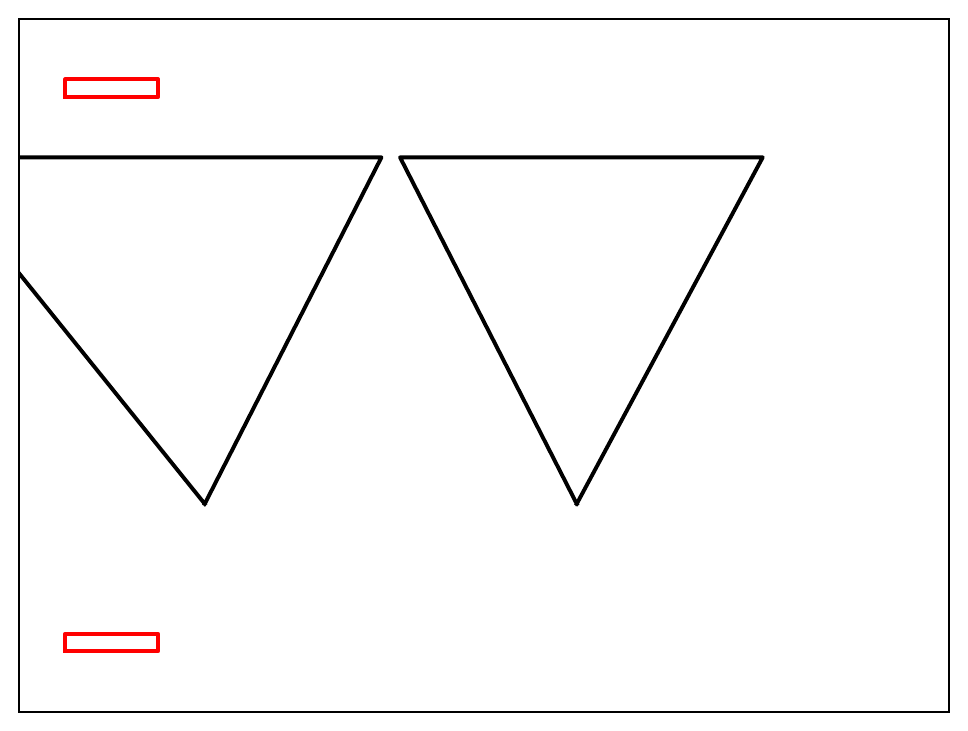}}
\hfill
\subfloat[TrapCup]{\label{fig:task4}\includegraphics[width=0.23\linewidth]{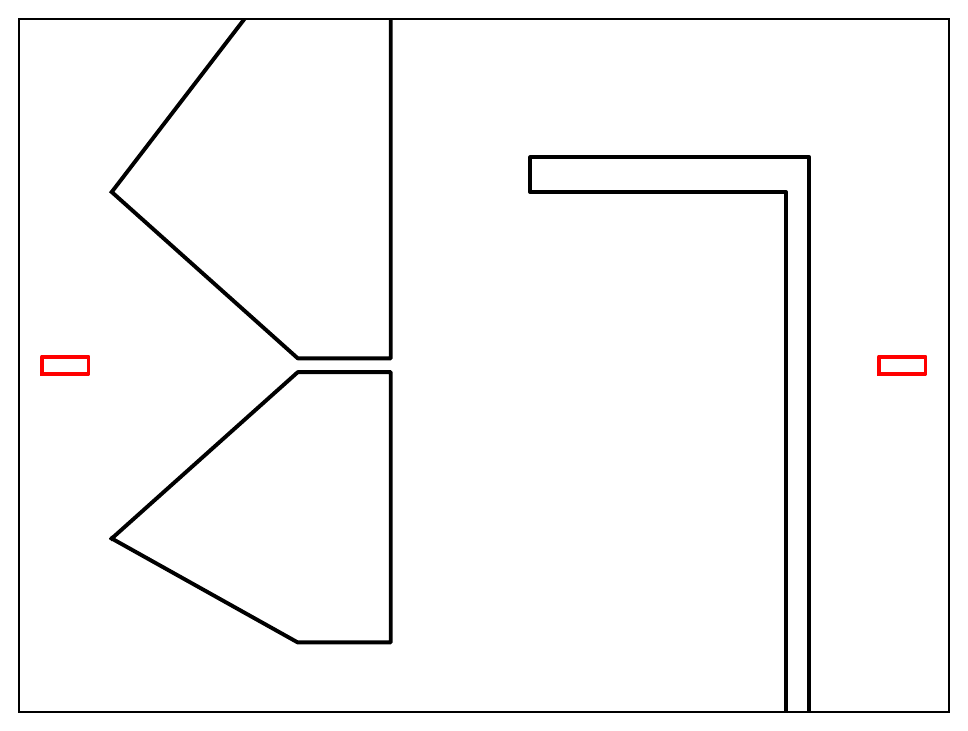}}
\caption{The environments in which we evaluate the different guiding space algorithms. We draw the robot in its start and goal configurations in red.}
\label{fig:tasks}
\end{figure*}

\subsection{Axes for Evaluating Guidance}
\label{sec:eval}

The metric described above measures the quality of each individual node expansion step of the guided search algorithm, but there are other options.

\subsubsection{Environment level} 
One can measure quality of guidance globally or in terms of local (possibly hierarchical) environment regions. In global evaluation we take an expectation of quality over tasks sampled from the environment. In local evaluation we subdivide the environment and compute our metric on each region independently, reporting the distribution of results over the full environment. Note that our method for global evaluation naturally applies to any subdivision of the environment, and we leave a discussion of appropriate environment partitioning to future work.

\subsubsection{Sample level} 
One can measure quality of guidance by considering both single and multiple samples. In single-step evaluation we measure the expected quality of a single node expansion of a single search tree. Such evaluation not only depends on the task and environment, but on the current exploration progress, namely information regarding previously sampled points in the configuration space encoded in the search tree. In multiple-sample evaluation we instead measure the holistic efficiency of all samples produced by the guiding space over the course of planning. 

In other words, single-step evaluation considers distributions over trees $T$, whereas multiple-sample evaluation considers distributions over $\C$, implicitly described by the distribution of valid trees produced by an algorithm. We refer the reader to \cite{attali2022evaluating} for an in depth discussion in this setting, and here summarize that in multiple-sample evaluation our metric reduces to measuring the entropy of the observed sampling distribution, where lower entropy means better guidance.



\begin{figure*}[t!]
\centering
\subfloat[RRT]{\label{fig:rrt_guiding_space}\includegraphics[scale=0.17]{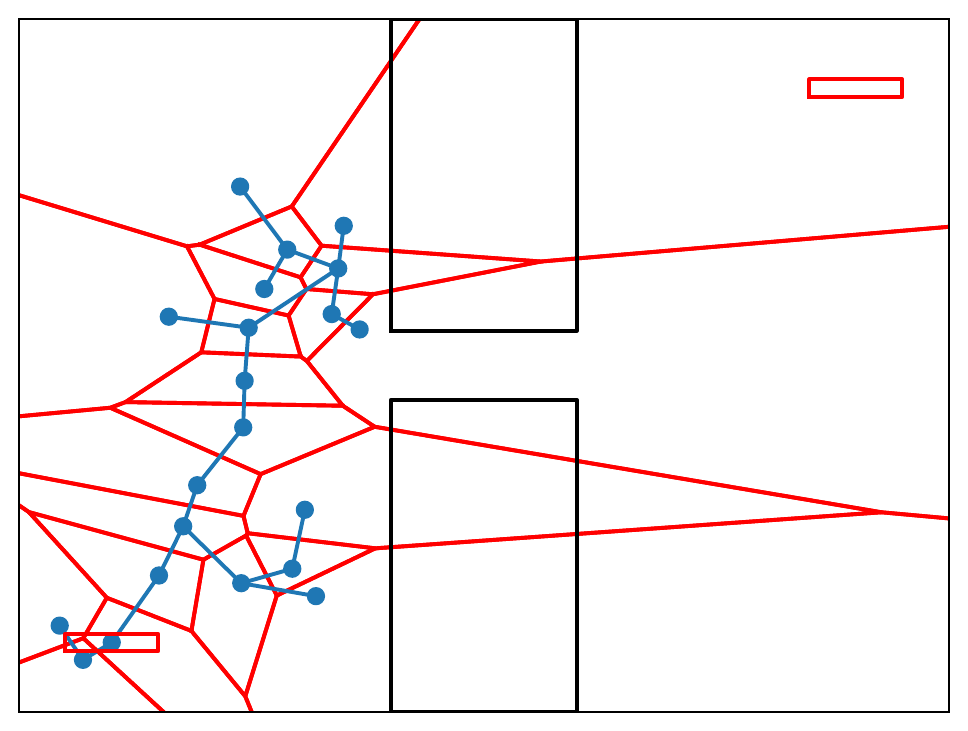}}
\hfill
\subfloat[LazyPRM]{\label{fig:lazyprm_guiding_space}\includegraphics[scale=0.17]{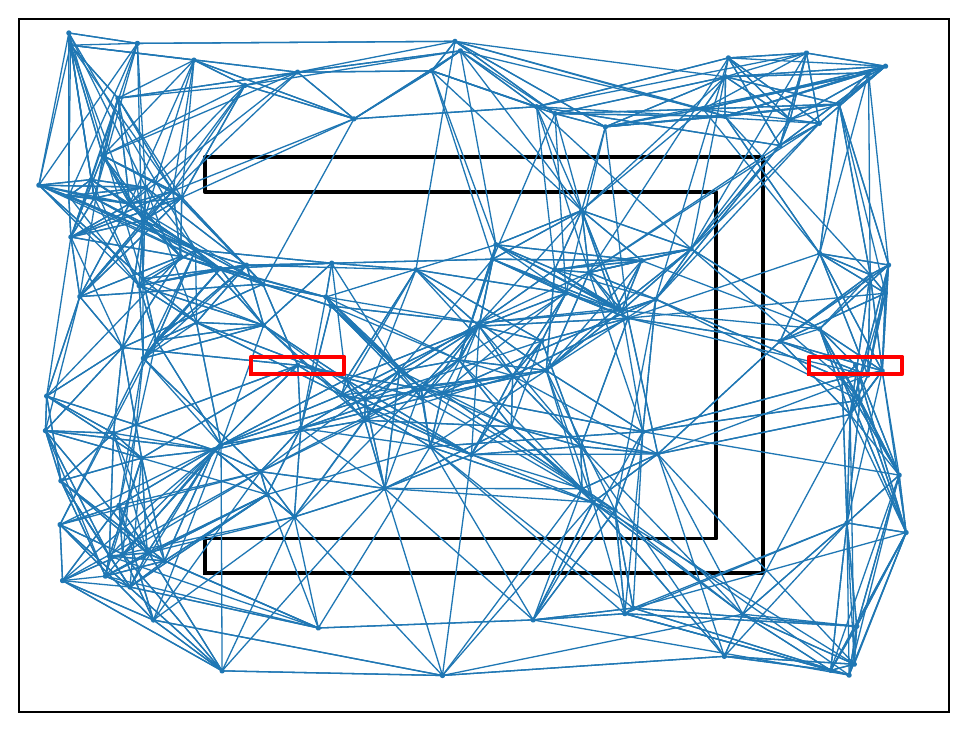}}
\hfill
\subfloat[MedialAxis]{
\label{fig:medialaxis_guiding_space}\includegraphics[scale=0.17]{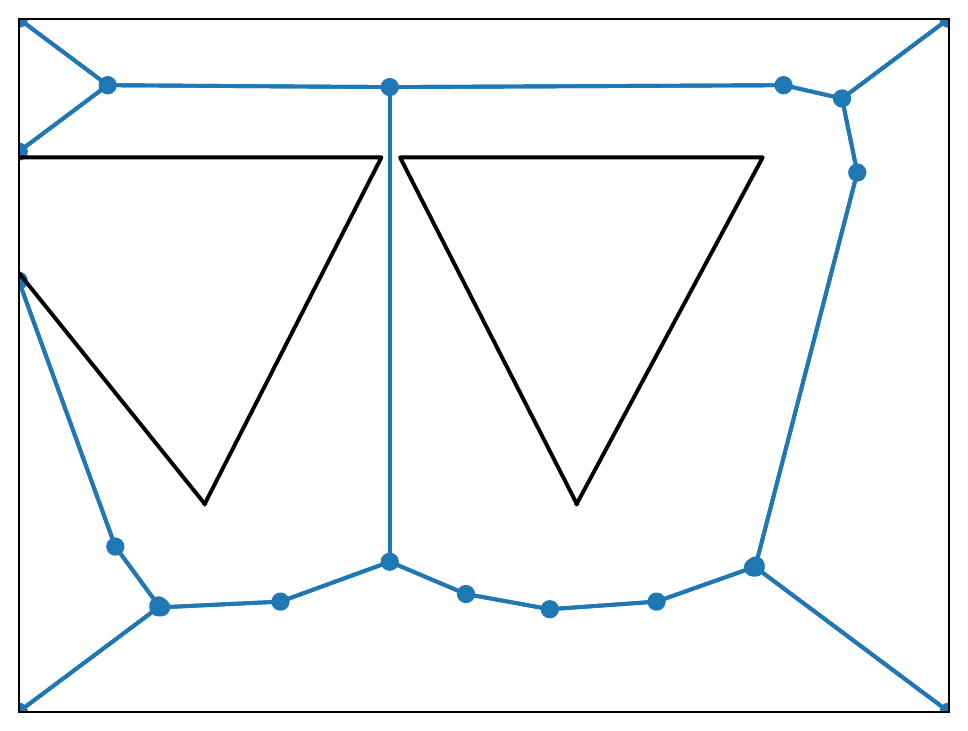}}
\hfill
\subfloat[PathDatabase]{\label{fig:lightning_guiding_space}\includegraphics[scale=0.17]{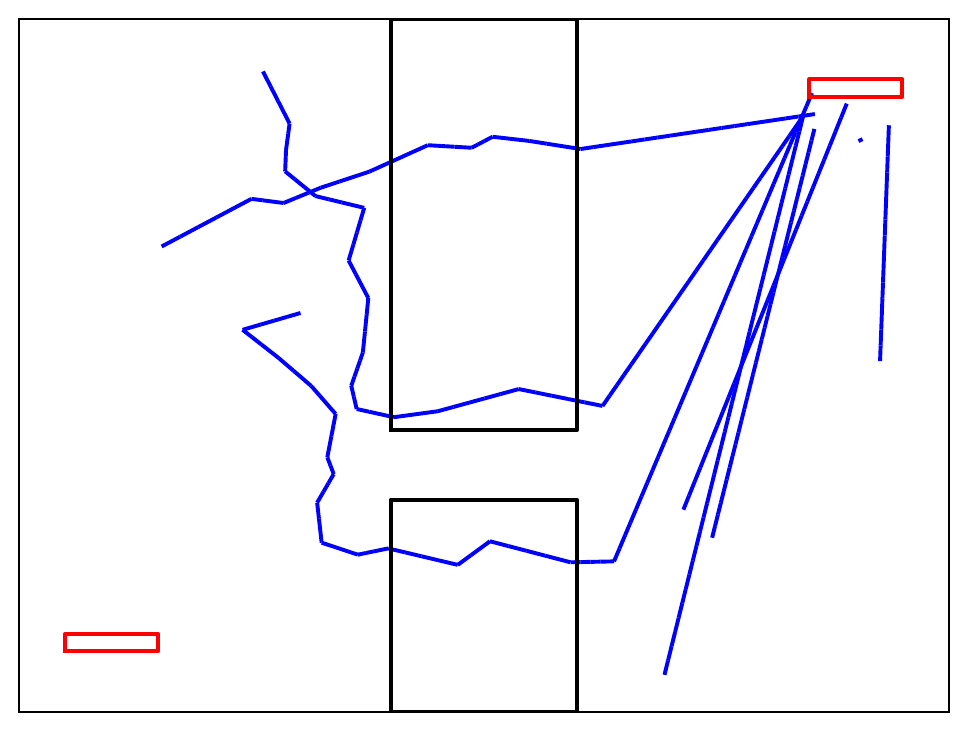}}
\caption{Example guiding spaces of each algorithm.
We display the starts and goals of the rectangular robot for (a) RRT, (b) LazyPRM, and (d) PathDatabase to indicate that these guiding spaces are in C-space, whereas (c) MedialAxis is in workspace.
(a) For RRT in SimplePassage, we show the Voronoi decomposition. 
(b) For LazyPRM in Cup, we show the unvalidated roadmap.
(c) For MedialAxis in Trap, we show the workspace skeleton.
(d) For PathDatabase in RandomSimplePassage, we display the paths from the database that pass near the goal.
}
\label{fig:guiding_spaces}
\end{figure*}

\begin{figure*}[t!]
\centering
\subfloat[RRT tree on SimplePassage]{\label{fig:RRT_tree}\includegraphics[scale=0.17]{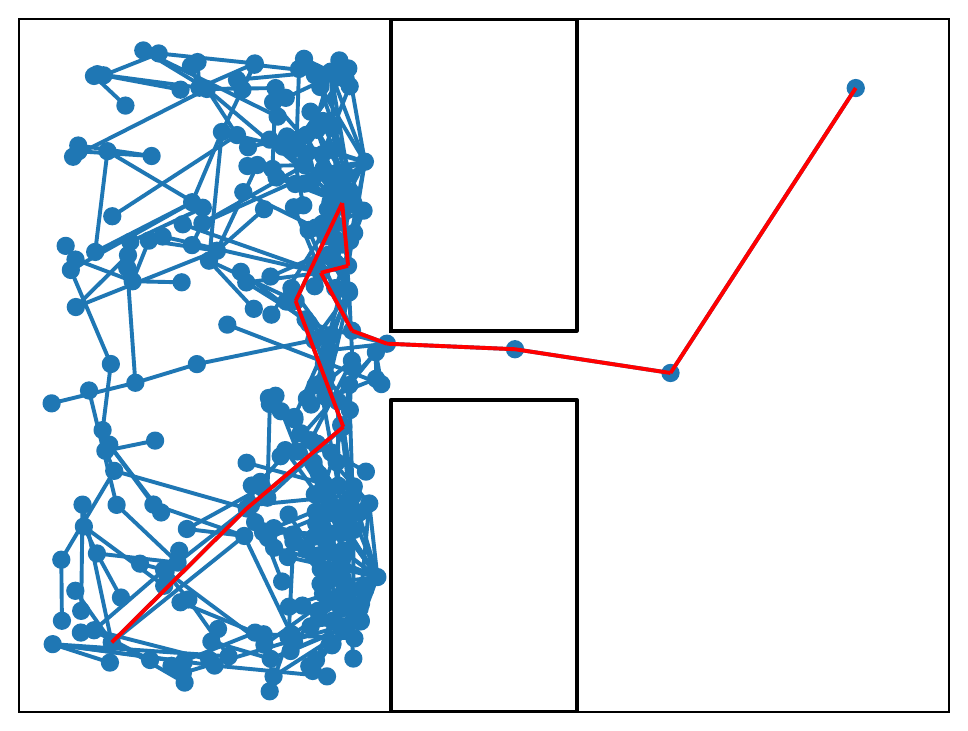}}
\hfill
\subfloat[LazyPRM tree on Cup]{\label{fig:Lazy_tree}\includegraphics[scale=0.17]{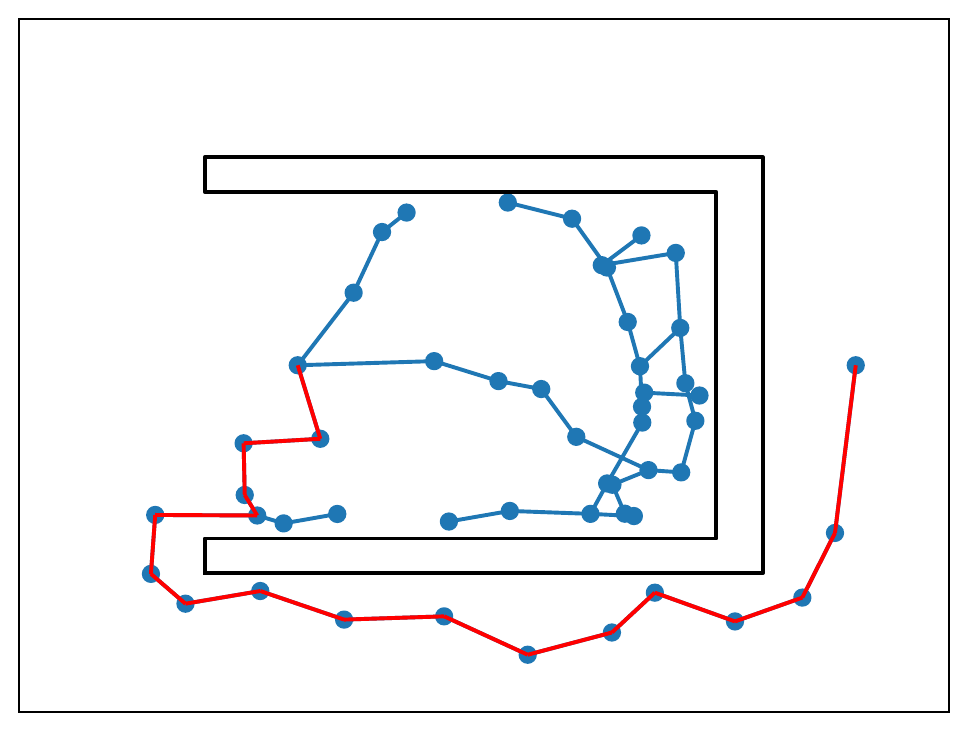}}
\hfill
\subfloat[MedialAxis tree on Trap]{\label{fig:MedialAxis_tree}\includegraphics[scale=0.17]{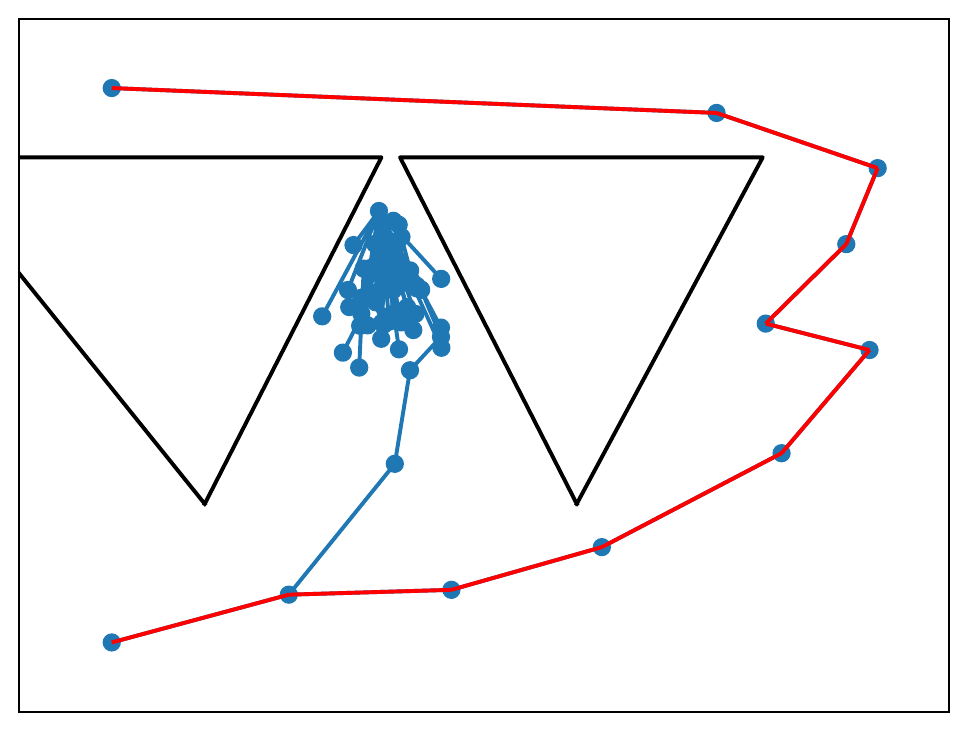}}
\hfill
\subfloat[PathDatabase tree on RandomSimplePassage]{\label{fig:Lightning_tree}\includegraphics[scale=0.17]{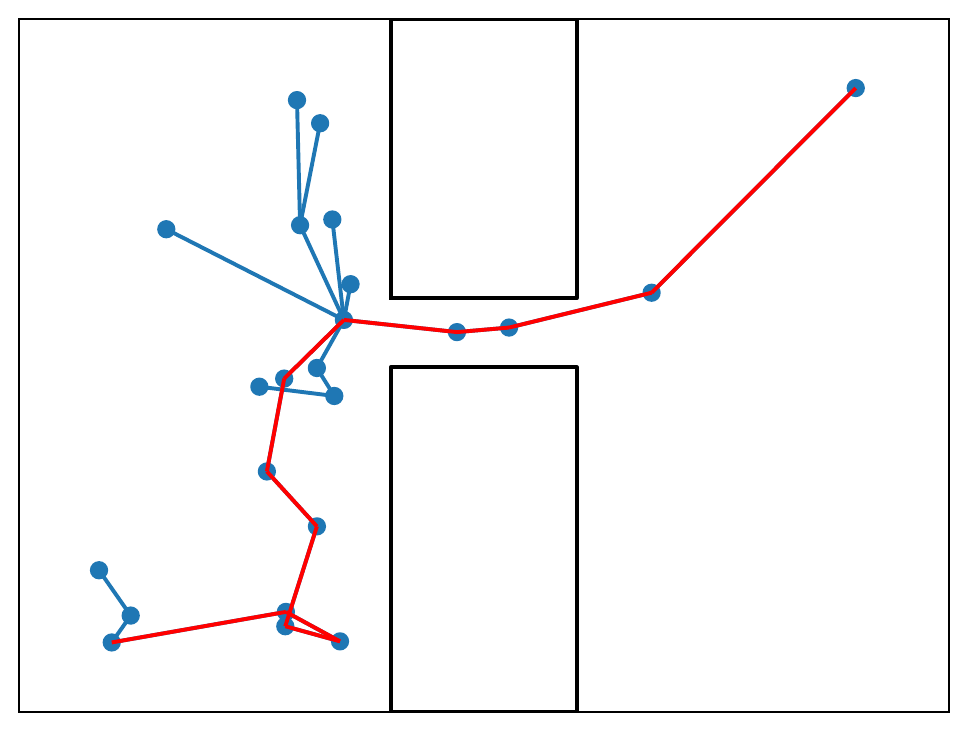}}
\caption{We show examples of each guidance method on the main environment it was tested on. Note that the visualizations are the 2D projections of the 3D C-space of a planar rectangular robot. In all cases, our implementation queries the local planner for valid edges to the goal, thus terminating the search once a configuration has been found that is visible from the goal. The final paths are highlighted in red.}
\label{fig:alg_examples}
\end{figure*}

\section{Refactoring Existing Algorithms}
\label{sec:refactoring}

Existing algorithms such as those cited in Section~\ref{sec:prior_work} are not implemented in the form of Algorithm~\ref{gs_alg}, rather each uses their respective guiding space in a different way. In the next section we conduct experiments on four guided search algorithms, RRT \cite{lavalle1998rapidly}, LazyPRM \cite{BohlinK00}, DR-RRT \cite{DennySBA16}, and Lightning \cite{berenson2012robot}. Figure~\ref{fig:alg_examples} shows examples of trees produced by these algorithms for solving the tasks depicted in Figure~\ref{fig:tasks}, and Figure~\ref{fig:guiding_spaces} visualizes their guiding spaces.

LazyPRM (environment modification) builds a probabilistic roadmap \emph{without} performing collision detection with the environment (or only doing so for vertices but not edges). Our implementation grows a tree using the length of the shortest path on this lazy graph as guidance for selecting which node to expand on the tree. When an invalid sample is found in C-space, we pass this information back to the guiding space (the lazy graph) by deleting nearby vertices/edges.

DR-RRT (robot modification) uses the medial axis skeleton computed in workspace to guide the robot. The original proposed implementation uses the edges of this skeleton to define ``dynamic regions'' along which to sample, and does so by creating balls around these regions and projecting back into C-space. For robots whose workspace configuration is a subset of the dimensions of their C-space this is easy, but for more complex C-space the general form of such a procedure involves an inverse-kinematics projection. Our new implementation, which we call MedialAxisGuidance is more straightforward - we project C-space nodes $v$ with forward-kinematics into workspace, then project these workspace poses onto the nearest medial axis edge $e_v$. The corresponding guidance value is then the distance along the skeleton edges from the projected edge to the projection of the goal, $d(e_v, e_t)$. Whenever a node $v$ fails to expand due to obstacle collision we increase the weight of $e_v$, thereby reducing the chance that the guidance encourages exploration along that edge.

Lightning (experience-based guidance) uses a database of paths from related motion planning problems to guide exploration. Namely, given a motion planning problem, a path that was produced by the most similar task (start and goal) is selected from the database, and then copied into the current C-space. Where the path is invalid the algorithm fixes the solution by running a bi-directional RRT from the two ends of the breakage. We highlight that this sub-problem can be just as difficult as the original, and moreover that the full power of the database is never used. Our new implementation, which we call PathDatabaseGuidance uses the \emph{length} of the queried path as the guidance value. As with LazyPRM, failed sample information is passed down to the guiding space by filtering out paths that pass close to invalid regions. If no node can locally plan to any path, then we default to RRT until a new node is found from which there is a visible path. We note that this implementation is relatively similar to \cite{aine2015learning}, but we query the database using task similarity as in \cite{berenson2012robot} rather than dynamic time warping, and more importantly we actively filter the database using information from the search tree.

\begin{figure*}[t!]
    \centering
    \subfloat[RRT evaluation on all four environments]{\label{fig:rrt_eval}\includegraphics[width=0.4\linewidth]{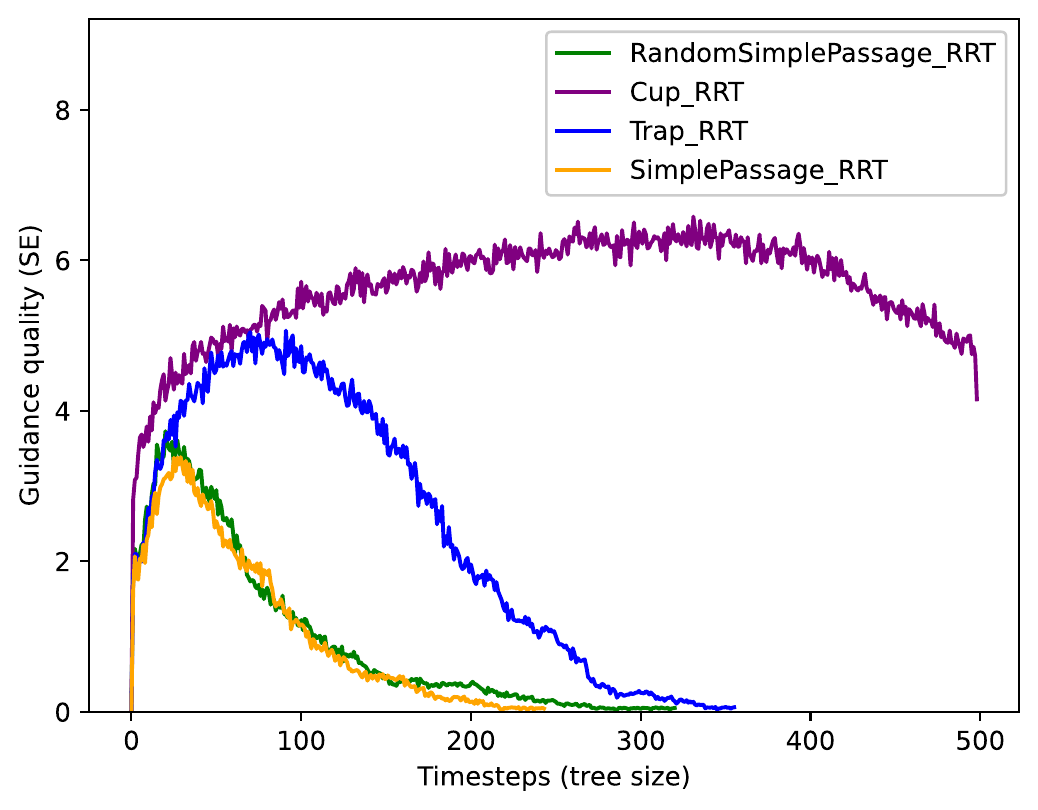}}
    \hspace{10mm}
    \subfloat[LazyPRM evaluation on Cup and SimplePassage]{\label{fig:lazy_eval}\includegraphics[width=0.4\linewidth]{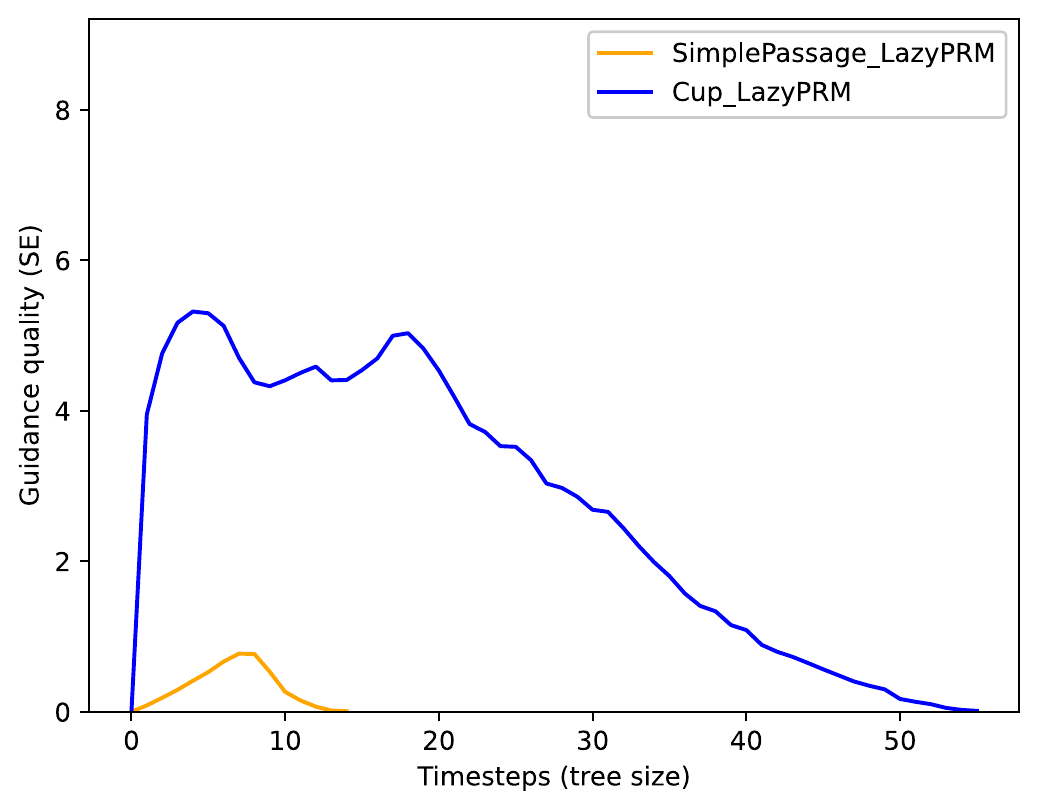}}
    
    \subfloat[MedialAxisGuidance evaluation on Trap and SimplePassage]{\label{fig:medialaxis_eval}\includegraphics[width=0.4\linewidth]{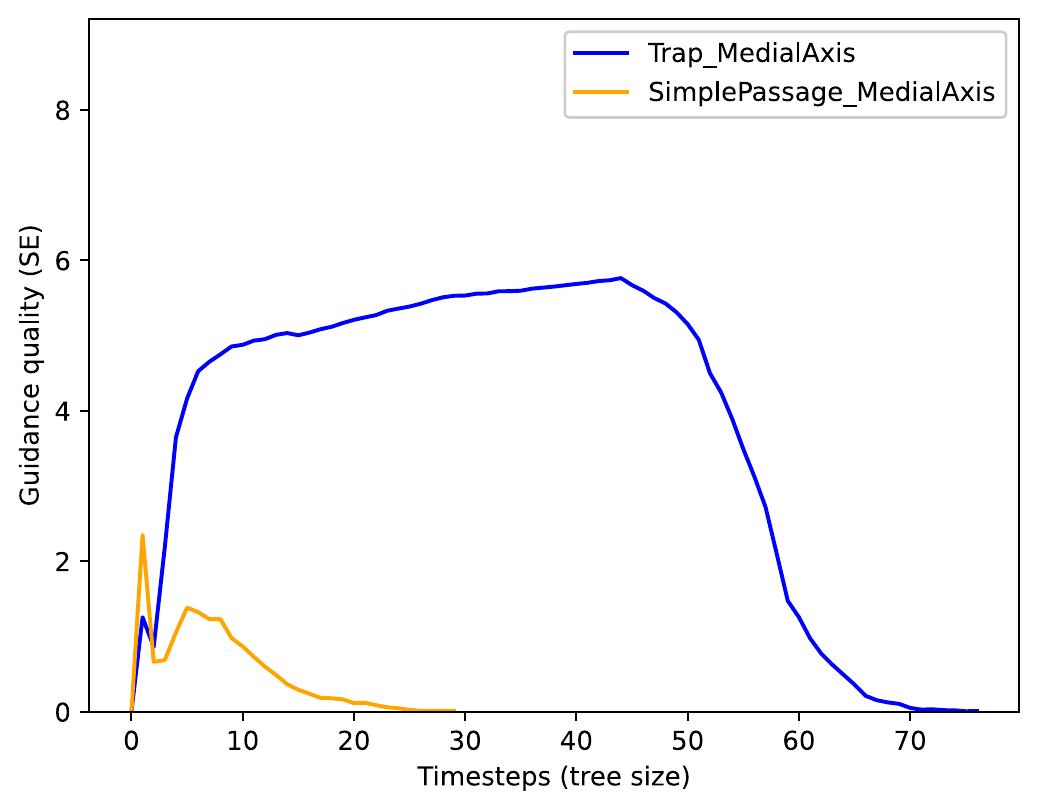}}
    \hspace{10mm}
    \subfloat[PathDatabase evaluation on the RandomSimplePassage environment with different size databases]{\label{fig:lightning_eval}\includegraphics[width=0.42\linewidth]{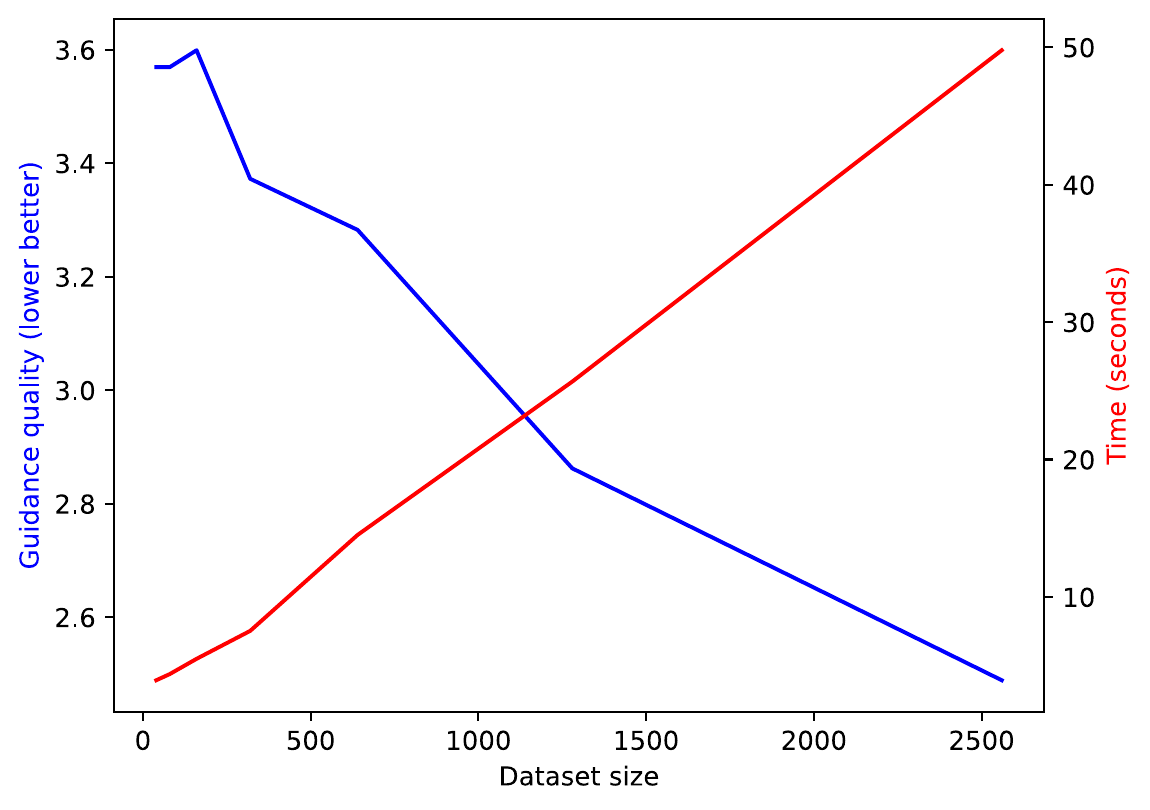}}
    \caption{RRT, LazyPRM, MedialAxis, and PathDatabase (Lightning) evaluations on TrapCup environment. All plots are averages over 128 random seeds. When a seed finishes the search we consider its sampling efficiency to be $0$ for all future samples (i.e., sampling the goal). Runtime in PathDatabase includes only search time and not database creation, and guidance quality is averaged over the entire run instead of shown per iteration as in the other plots.}
    \label{algorithm_evals}
\end{figure*}

\section{Experiments}
 \label{sec:experiments}

\begin{figure*}[b!]
\centering
\subfloat[TrapCup guidance over time]{\label{fig:trapcup_eval}\includegraphics[width=0.4\linewidth]{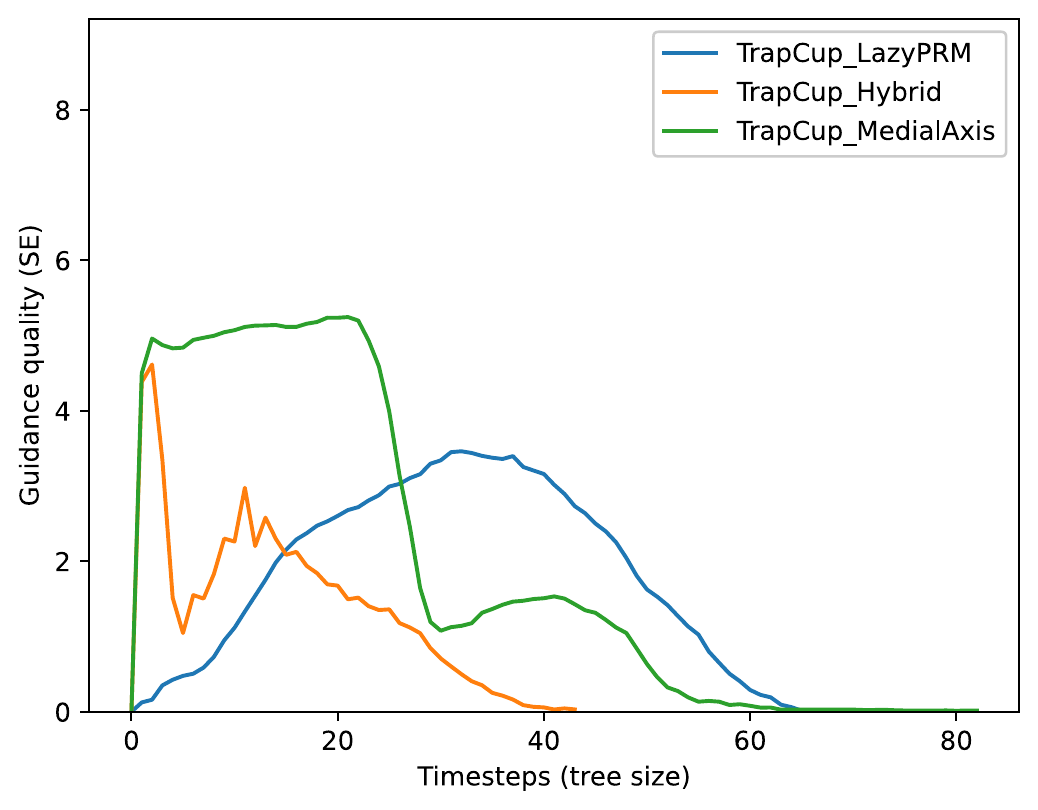}}
\hspace{10mm}
\subfloat[MedialAxis search tree]{
\label{fig:trapcup_medialaxis}\includegraphics[width=0.4\linewidth]{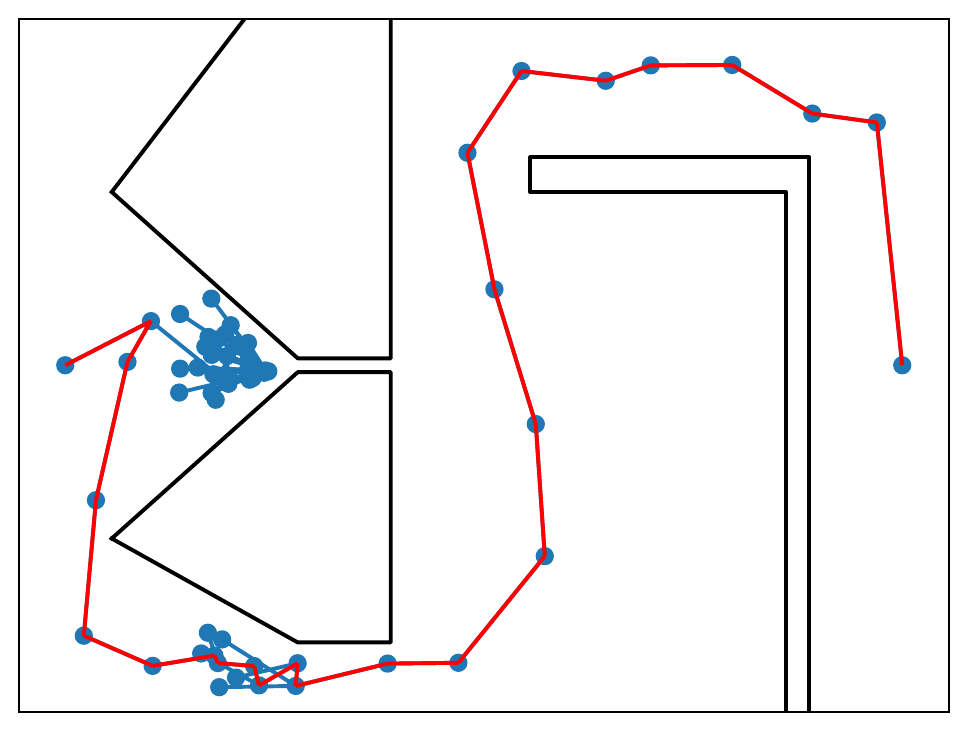}}

\subfloat[LazyPRM search tree]{\label{fig:trapcup_lazy}\includegraphics[width=0.4\linewidth]{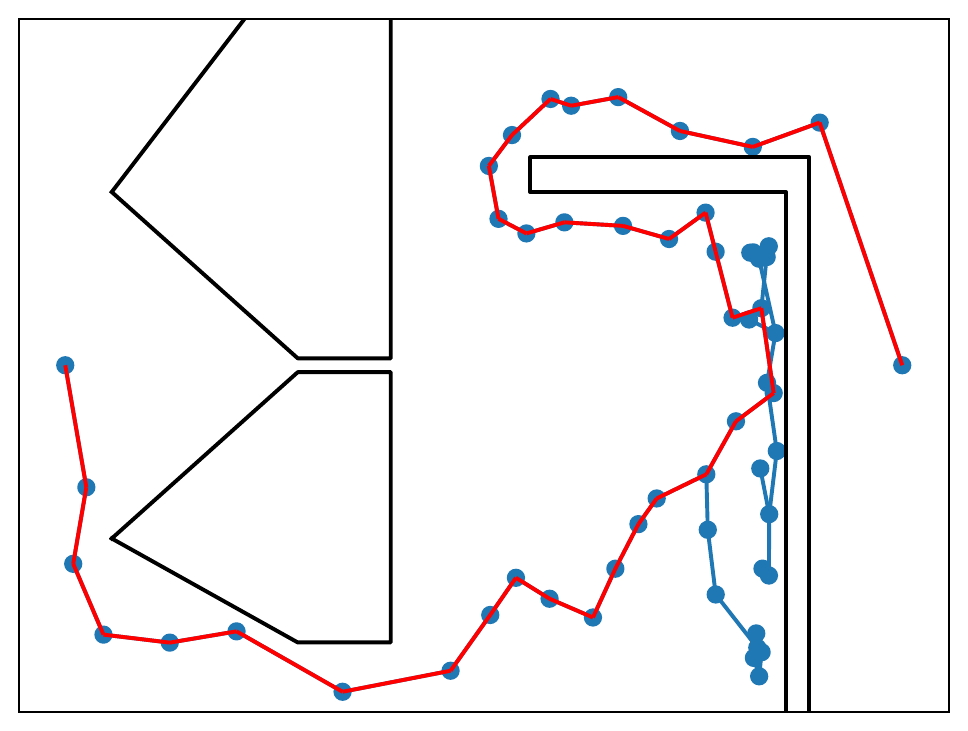}}
\hspace{10mm}
\subfloat[Hybrid method search tree]{\label{fig:trapcup_multi}\includegraphics[width=0.4\linewidth]{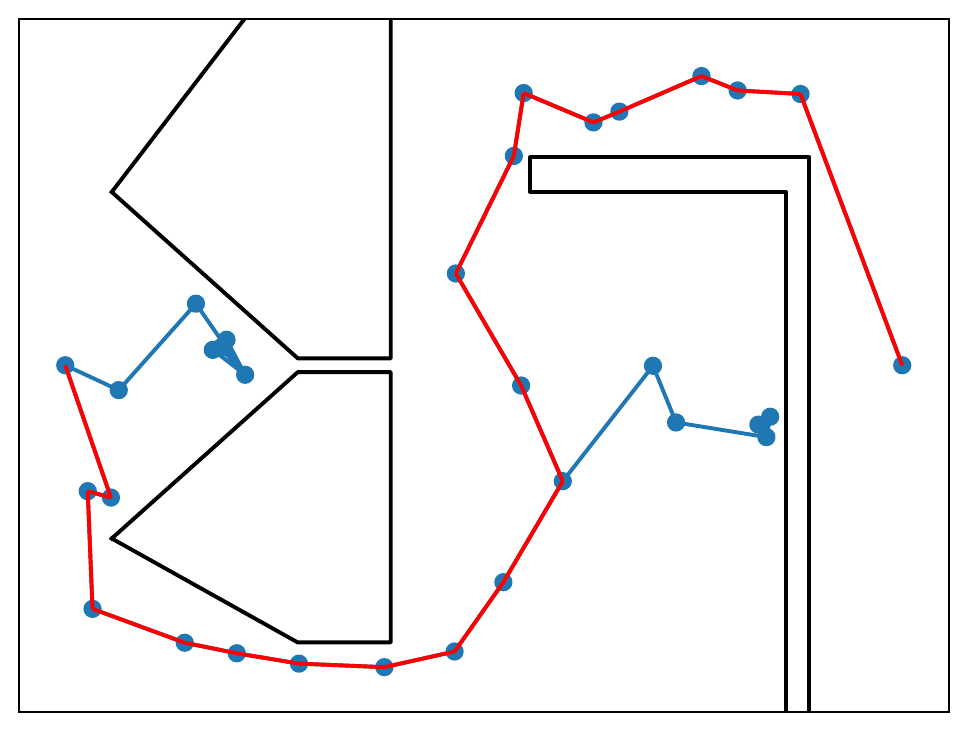}}
\caption{Results of the TrapCup environment experiments in which the rectangular mobile robot must navigate from the center-left to the center-right. MedialAxis struggles with the local minima of the first obstacle, which resembles the Trap environment. LazyPRM struggles with the local minima of the second obstacle, which resembles the Cup environment. A hybrid guiding space which uses both projections can escape both minima faster and has overall better guidance.}
\label{fig:trapcup}
\end{figure*}

We design environments intended to demonstrate both strengths and weaknesses of each guiding space mentioned in Section~\ref{sec:refactoring}, using a rectangular mobile base (3-dimensional C-space) in a 2-dimensional workspace. SimplePassage has two rooms connected by a narrow passage, Cup has a narrow ``C''-shaped obstacle that creates a dead-end for search, and finally Trap contains two passages where the first is not valid in C-space. The evaluations are done with a smoothed target distribution that is at least $\epsilon = 0.0001$ at each node on the tree. In addition, $\tau_v, \delta_v$ values were normalized and we set $\tau=\delta=0.1$.

We show results in Figure~\ref{algorithm_evals}, plotting quality of guidance (lower is better) over the course of search progress, i.e., number of samples. RRT initially performs similarly on all environments, doing well and then poorly as it struggles to get through narrow passages. LazyPRM performs well on SimplePassage but poorly on Cup, where it spends time exploring the dead end. MedialAxisGuidance does well on SimplePassage but poorly on Trap, where a valid passage in workspace is not valid in C-space. These first three experiments demonstrate that guidance matches our intuition for when each algorithm performs well or struggles, which can be seen qualitatively in Figure~\ref{fig:alg_examples}. In addition, we highlight that RRT is comparable to other algorithms initially, and especially on environments where MedialAxisGuidance or LazyPRM lead the search astray. 

Next we test PathDatabaseGuidance on a dataset of RandomSimplePassages, an environment with a randomly placed narrow passage between two rooms. We show how guidance improves as dataset size increases, while runtime gets worse (as querying the database takes longer), showing how such traditional metrics can obscure properties of algorithms because they are implementation dependent. This matches the intuition mentioned in Section~\ref{sec:simple_examples} regarding the trade-off between guidance quality and the difficulty of computing it. 

\subsection{Multi-Guidance Search}
\label{multiguidance}

Finally, we create a hybrid environment TrapCup which combines features from both the Trap and Cup environments: first the robot must avoid a passage that is valid in workspace but not C-space, then it must avoid a dead-end region. We now design a hybrid guiding space that is a combination of LazyPRM and MedialAxis. In other words, it creates both guiding spaces (i.e., the cross product) and uses a rule to determine how to select or combine the heuristics from the two. In the experiments shown in Figure~\ref{fig:trapcup} we simply swap the source of the guidance each time the search collides with an obstacle. Intuitively, each form of guidance faces certain local minima and by using multiple guiding spaces we can quickly escape those minima by passing control to another guiding space. Notice that the individual guidance plots for each guiding space indicate that these local minima are distinct (e.g., the curves have different peaks), and thus we see the hybrid guiding space out-performs its parts. Moreover, using number of samples alone as a metric would obscure this fact since MedialAxis and LazyPRM take about the same number of samples to find the goal. These results parallel what one would expect from running a multi-heuristic A* search \cite{aine2016multi} as opposed to single heuristic search, and highlights the value of framing sampling based motion planning as guided search; we now have a structured way of combining different algorithmic ideas from the literature as multiple guiding spaces, whereas the original implementations are not so conducive to modularization.

\section{Conclusion}

In this work we presented the guiding space framework, in which an explicit guiding space provides heuristic guidance to a tree based search algorithm in a configuration space. We showed how many existing methods fit this framework, albeit implicitly. By viewing such methods as guiding spaces, we can more easily compare and combine them into hybrid algorithms. As such, we defined \emph{sampling efficiency} as a way to measure the quality of biased node selections of a search tree, and showed how this metric matches intuition for a variety of algorithms and environments.

\bibliography{refs}
\end{document}